\documentclass[10pt,twocolumn,letterpaper]{article}
\usepackage{cvpr}

\usepackage{times}
\usepackage{graphicx}
\usepackage{amsmath}
\usepackage{amssymb}
\usepackage{multirow}
\usepackage{booktabs}
\usepackage{tabularx}
\usepackage[dvipsnames,table,xcdraw]{xcolor}
\usepackage{overpic}
\usepackage{colortbl}
\usepackage{makecell}
\usepackage{contour}
\usepackage[accsupp]{axessibility}
\usepackage[pagebackref=false,breaklinks,colorlinks,citecolor=RoyalBlue,bookmarks=false]{hyperref}

\usepackage{enumitem}
\usepackage{pifont}

\newcommand{\myParaP}[1]{\vspace{.05in}\noindent\textbf{#1}}
\newcommand{\myPara}[1]{\vspace{.05in}\noindent\textbf{#1:}}

\newcommand{\highlight}[1]{\textbf{\textcolor{ForestGreen}{#1}}}

\def\cvprPaperID{1567} 
\def\confYear{CVPR 2022}

\usepackage[capitalize]{cleveref}
\crefname{section}{Sec.}{Secs.}
\Crefname{section}{Section}{Sections}
\Crefname{table}{Table}{Tables}
\crefname{table}{Tab.}{Tabs.}

\newcommand{\figref}[1]{Fig.~\ref{#1}}
\newcommand{\tabref}[1]{Tab.~\ref{#1}}
\newcommand{\secref}[1]{Sec.~\ref{#1}}

\newcommand{\eqnref}[1]{Eqn.~(\ref{#1})}

\graphicspath{{../figure/}, {../figure/Related/}}

\begin{document}

\title{L2G: A Simple Local-to-Global Knowledge Transfer Framework for \\ Weakly Supervised Semantic Segmentation}

\author{
  Peng-Tao Jiang$^1$ ~~ Yuqi Yang$^1$ ~~ Qibin Hou$^1$\thanks{Qibin Hou is the corresponding author.}  ~~ Yunchao Wei$^2$ \\
$^1$TMCC, CS, Nankai University \quad
$^2$Beijing Jiaotong University \\
{\tt\small pt.jiang@mail.nankai.edu.cn  
andrewhoux@gmail.com }}

\maketitle

\begin{abstract}
Mining precise class-aware attention maps, a.k.a, class activation maps, 
is essential for weakly supervised semantic segmentation.
In this paper, we present L2G, a simple online local-to-global 
knowledge transfer framework for high-quality object attention mining.
We observe that classification models can discover 
object regions with more details 
when replacing the input image with its local patches.
Taking this into account, we first leverage a local classification network
to extract attentions from multiple local patches randomly cropped from
the input image.
Then, we utilize a global network to learn complementary 
attention knowledge across multiple local attention maps online.
Our framework conducts the global network to learn the captured
rich object detail knowledge from a global view and thereby
produces high-quality attention maps that can be directly 
used as pseudo annotations for semantic segmentation networks.
Experiments show that our method attains 72.1\% and 44.2\% mIoU scores
on the validation set of PASCAL VOC 2012 and MS COCO 2014, respectively, 
setting new state-of-the-art records.
Code is available at \url{https://github.com/PengtaoJiang/L2G}.
\end{abstract}

\section{Introduction} \label{sec:intro}
Deep learning algorithms \cite{long2015fully,lin2016refinenet,zhao2016pyramid} 
have promoted the rapid development of the semantic segmentation task 
in recent years. 
However, training a deep neural network for semantic segmentation requires 
a large number of pixel-wise accurate labels, which consume lots of human labors
and resources.
Recently, to reduce the reliance on accurate annotations,
researchers have attempted to study semantic segmentation 
based on cheap supervisions, 
such as bounding boxes \cite{papandreou2015weakly,dai2015boxsup}, 
scribbles \cite{lin2016scribblesup,vernaza2017learning}, points \cite{bearman2016s}, 
and image-level labels \cite{wei2017object,hou2016mining}.
Among these weak supervisions, image-level labels only provide information on
the existence of the target object categories, making them more popular
than other supervisions due to the easy way to collect.
In this paper, we also focus on weakly supervised semantic segmentation (WSSS) 
based on image-level labels.

\begin{figure}[t]
    \centering
    \setlength\tabcolsep{1pt}
    \begin{overpic}[width=0.48\textwidth]{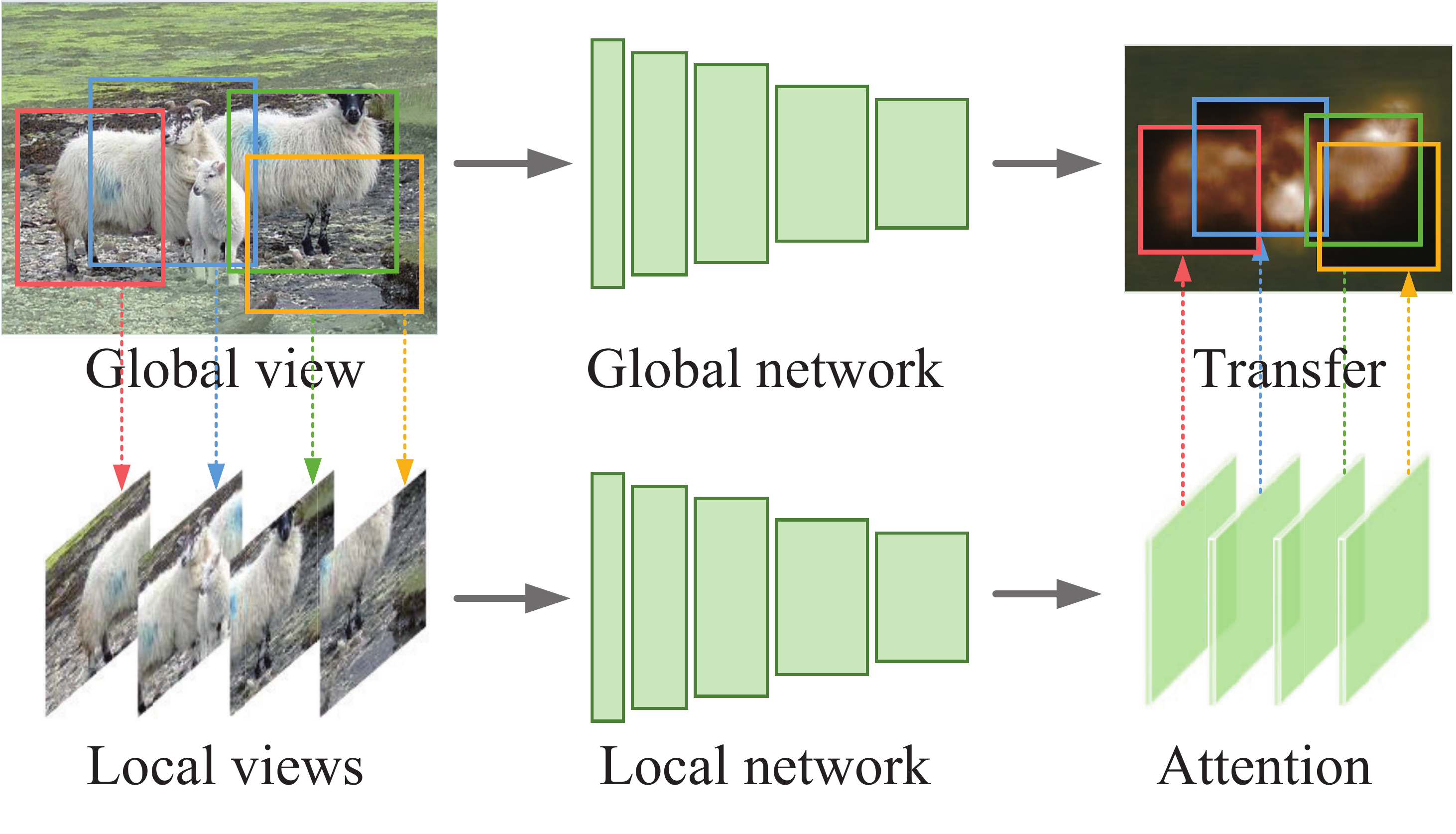}
    \end{overpic}
    \caption{Conceptual working pipeline of the proposed method. 
    We utilize the attention maps for local views with rich details
    extracted from the local network  to teach the global network.
    This enables the global network to learn the rich local details knowledge from the
    local network online and thereby more integral object attentions.
    }\label{fig:teaser}
    \vspace{-1pt}
\end{figure} 

Speaking of WSSS, one of the most important components should be the class activation map (CAM)~\cite{zhou2016learning} which contains both semantic and location
information about the target objects and can be used as pseudo pixel-level annotations 
for training segmentation networks.
Since the quality of CAMs has a great influence on the segmentation results, recently, 
many strategies have been proposed to advance the original CAM method, including 
adversarial erasing~\cite{wei2017object,zhang2018adversarial,hou2018self,zhou2020multi}, 
online attention accumulation~\cite{jiang2019integral,jiang2021online}, 
seed region expansion~\cite{kolesnikov2016seed,huang2018weakly}, 
and affinity learning~\cite{ahn2018learning,ahn2019weakly,xu2021leveraging}, \etc.
Despite the good performance, these works mostly take the whole input image as
the sole input to the model.
However, we empirically observe that classification models can 
discover more discriminative regions when taking local image patches as 
input compared to the whole input image.
This suggests a proper way to improve the quality of
attention maps by making use of local image patches.

In this paper, taking the above analysis into account, we present a simple
online local-to-global knowledge transfer framework, termed L2G, for generating 
high-quality object attentions.
%
%
A conceptual illustration has been depicted in \figref{fig:teaser}.
Different from the aforementioned attention mining 
strategies, we propose to take advantage of both the global view and 
the local views randomly cropped from the input image (regions enclosed by the colorful bounding boxes).
Specifically, our framework contains a local network that produces
local attentions with rich object details for local views as well as 
a global network that receives the global view as input and aims to 
distill the discriminative attention knowledge from the local network.

Our method offers the following advantages.
First of all, we produce attention maps from multiple local views of
the input image rather than its global view.
This allows us to attain more details on undiscovered semantic regions,
which are also complementary across different local views,
as shown in Fig.~\ref{fig:motivation}.
%
%
Second, by designing a knowledge transfer loss, the complementary attention knowledge
can be efficiently transferred to the global network in an online learning manner.
This enables the global network to capture pixel-level semantic object details and produce high-quality attention maps in inference.
%
Last but not the least, the overall pipeline is simple and flexible. We can selectively add 
additional constraints~\cite{lee2021railroad} to the local network 
to help shape the attained object attentions.

\newcommand{\addFig}[1]{\includegraphics[width=0.195\linewidth]{figure/motivation/#1}}
\newcommand{\addFigs}[1]{\addFig{#1_0.jpg} & \addFig{#1_2.jpg} & \addFig{#1_1.jpg} & \addFig{#1_3.jpg} & \addFig{#1_4.jpg}}

\begin{figure}[t] 
\centering
\small
\setlength\tabcolsep{0.6pt}
\renewcommand{\arraystretch}{0.7}
\begin{tabular}{ccccc}
  \addFigs{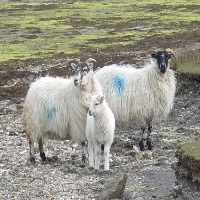} \\
  \addFigs{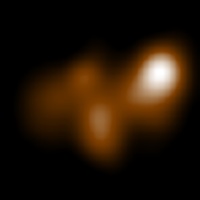} \\
\end{tabular}
\caption{Motivation of our L2G attention knowledge transfer method.
  The top row shows the original image (global view) and multiple image patches 
  after random crop (local views). 
  The second row shows the attention maps generated by CAM \cite{zhou2016learning}.
  We can observe that the attention maps of the local views capture more object
  details compared to that of the global view.
}
\label{fig:motivation}
\vspace{-10pt}
\end{figure}

We evaluate our method on the PASCAL VOC 2012 and 
MS COCO 2014 datasets. 
Experiments demonstrate that our method achieves better 
performance than previous state-of-the-art methods.
When using the DeepLab-v2 model \cite{chen2017deeplab} as our segmentation network,
we attain 72.1\% and 71.7\% mIoU scores on the
validation set and the test set of PASCAL VOC 2012, 
and 44.2\% on the validation set of MS COCO 2014,
setting new state-of-the-art records under the weakly supervised setting.
We also conduct a series of ablation experiments to help readers better 
understand how each component performs in our method.
%

\section{Related Work}

\subsection{Weakly Supervised Semantic Segmentation}

\myParaP{One-stage WSSS methods} directly utilize the image-level labels 
as supervision to train an end-to-end segmentation network.
Early works \cite{pathak2014fully,pinheiro2015image} formulate this problem 
as multiple instance learning.
Later, Papandreou \etal \cite{papandreou2015weakly} proposed an 
Expectation-Maximization (EM) method that utilizes 
the intermediate prediction to supervise the segmentation network.
%
%
Zhang \etal \cite{zhang2020reliability} utilized the image 
classification branch to generate attention maps and constructed 
the pseudo segmentation labels to supervise the parallel segmentation branch.
Araslanov \etal \cite{araslanov2020single} proposed a self-supervised 
mechanism that applies the image appearance priors to generate pseudo 
segmentation labels during training.
Chen \etal \cite{chen2021end} constructed an end-to-end framework 
that uses an encoder-decoder network to explore object boundaries.
Compared to two-stage WSSS methods, one-stage methods usually 
have inferior performance and are 
less attractive.

\myParaP{Two-stage WSSS methods} rely on attention maps 
\cite{zhou2016learning,zhang2018adversarial} 
to generate pseudo segmentation labels, which are then used 
to train segmentation networks.
%
%
The core of two-stage WSSS methods is to produce high-quality attention maps
\cite{wei2018revisiting,chang2020mixup,wang2020self,sun2020mining,li2018tell}.
Towards this goal, a lot of works have been proposed recently.
Wei \etal \cite{wei2017object} proposed the adversarial erasing strategy, 
which iteratively occludes the mined object regions
to drive the classification network to discover new object regions. 
Hou \etal \cite{hou2018self} improved the adversarial erasing strategy 
by using a self-erasing strategy to prevent attention 
from spreading to the background.
Kolesnikov \etal \cite{kolesnikov2016seed} introduced the seed-expansion idea, 
which expands the initial seed regions from the pre-computed attention maps 
and constrains the expanded regions to align with the object boundaries.
%
%
Later, Jiang \etal \cite{jiang2019integral} proposed the online 
attention accumulation strategy that utilizes the attention maps 
of different training phases.
Chang \etal \cite{chang2020weakly} exploited the sub-category information 
to highlight the non-discriminative semantic regions.

\begin{figure*}[t]
    \centering
    \setlength\tabcolsep{1pt}
    \begin{overpic}[width=\textwidth]{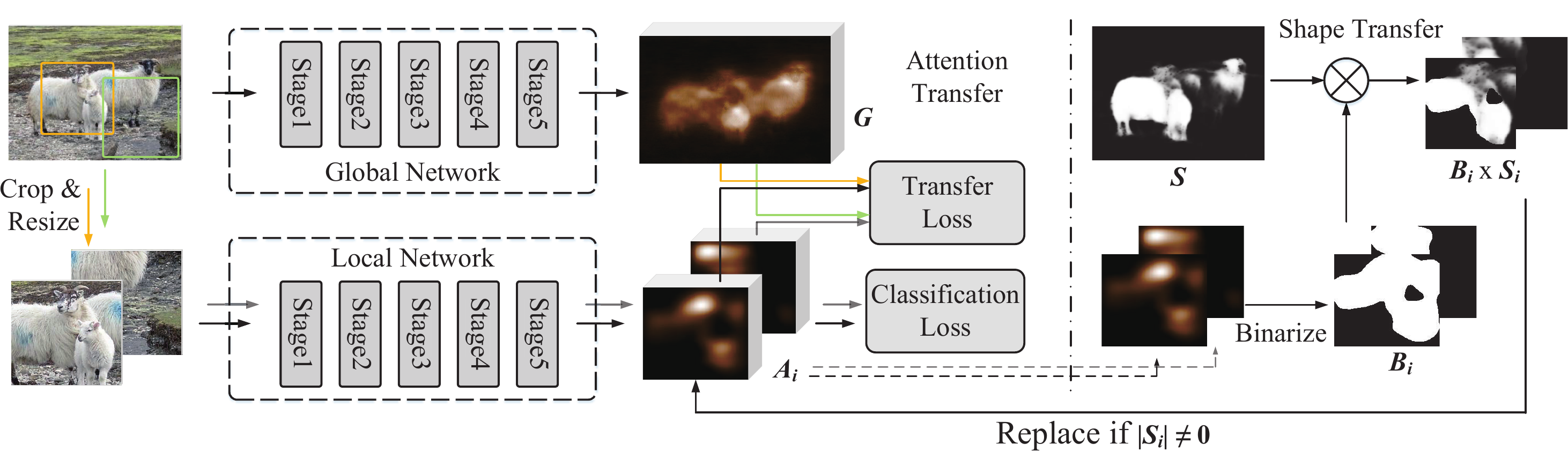}
    \end{overpic}
    \caption{Overall framework of the proposed method. The complementary attention
    maps captured by the local network is distilled into the global network
    by a knowledge transfer loss.
    }\label{fig:pipeline}
\end{figure*} 

Another line of works attempts to refine attention maps 
to obtain integral object regions with precious boundaries.
Ahn \etal \cite{ahn2018learning} learned pixel affinity 
to propagate the semantics of strong responses in attention maps 
to the adjacent pixels.
Chen \etal \cite{chen2020weakly} and Ahn \etal \cite{ahn2019weakly} 
further improved this method by explicitly learning the class boundaries.
Lee \etal \cite{lee2021railroad} utilized the off-the-shelf 
saliency maps as supervision to guide the region learning 
to generate high-quality attention maps.

One common point shared by the aforementioned methods is that 
they all refine attention maps on the image's global view.
Differently, our method takes advantage of both the global view and
multiple local views and studies how to efficiently transfer 
the complementary attention knowledge from the local network to the global network
to improve the quality of attention maps.
%

\subsection{Knowledge distillation}
Our work is also related to knowledge distillation \cite{hinton2015distilling,furlanello2018born,yuan2020revisiting}, 
which aims to distill the knowledge from the well-trained teacher model 
to a student model.
For the image classification task, these works focus on improving
the student model by imitating the prediction distribution of
the teacher model.
Moreover, some researchers \cite{liu2019structured,he2019knowledge} 
also study knowledge distillation for the semantic segmentation task.
%
%
%
Differently, we investigate how to transfer the attention knowledge captured 
by the local views to the global network in an online learning manner
to better leverage the complementary information from multiple views.

\section{Method}

In this section, we present the whole framework of our method in detail.
Before describing the framework, we first give some fundamental 
introduction to attention map generation.
%
%

\subsection{Prerequisites}
We first present the way to generate attention maps.
%
%
Given an input image $I$, let $y$ be the image-level label.
The output feature $F$ of the last convolutional layer has $C$ channels, 
identical to the number of classes.  
The last convolutional layer is followed by a global average pooling layer, 
where the feature $F$ is pooled to a vector $f^{C}$ of size $C$.
We calculate the classification loss by applying a sigmoid cross-entropy loss function, 
which is formulated as follows: 
\begin{equation} \label{eqn:ce}
    L_{\text{ce}} = - \frac{1}{C} \sum_{c=1}^C y^c \log(\sigma(f^c)) + (1-y^c) \log(1 - \sigma(f^c)),   
\end{equation}
where $\sigma$ is the sigmoid function.
The attention maps can be generated from the output of the last convolutional layer.
For some class $c$, the attention map $A^c$ is derived from the $c^{th}$ channel of $F$, 
which can be formulated as
\begin{equation} \label{eqn:cam}
A^c = \frac{\mbox{ReLU}(F^c)}{\max(\mbox{ReLU}(F^c))}.
\end{equation}

The above method, as pointed out in most previous work 
\cite{wei2017object,hou2018self,jiang2019integral,ahn2018learning},
can only locate the most discriminative regions.
It often fails in discovering those non-discriminative 
object regions that are semantically meaningful as well.
%
%
In the following, we propose a novel attention generation framework by
presenting a new local-to-global knowledge transfer method to
capture high-quality object attentions.

%

\subsection{Overall Framework}

As mentioned in \secref{sec:intro}, the local network focusing on
processing local patch views tends to discover more discriminative object regions.
Based on this observation, we propose to leverage the attention maps 
for local views to aid a global network to locate more integral 
object regions.

The overall framework of the proposed approach can be found 
in \figref{fig:pipeline}.
Functionally, there are four components: a global network, a local
network, an attention transfer module, and a shape transfer module.
The global network and the local network can be any CNN classifier,
such as the popular VGGNet~\cite{simonyan2014very} 
or ResNet-38~\cite{wu2019wider}.
In the attention transfer module, we optimize two loss functions:
a classification loss $L_{\text{cls}}$ that is used to recognize the
semantic objects and an attention transfer loss $L_{\text{at}}$
that encourages the global network to imitate the local network
to discover more discriminative regions.
In the shape transfer module, we introduce a shape constraint to
loss $L_{\text{at}}$, yielding $L_{\text{st}}$,
to shape the captured object attentions.
Therefore, the overall optimized loss function can be formulated as follows:
\begin{equation}
    L = L_{\text{cls}} + \lambda \cdot L_{\text{kt}}, 
\end{equation}
where $\lambda$ denotes the loss weight for $ L_{\text{kt}}$.
When no shape constraint is added, $L_{\text{kt}} = L_{\text{at}}$.
Otherwise, $L_{\text{kt}} = L_{\text{st}}$.

\subsection{Local-to-Global Attention Transfer} \label{sec:ATr}

Given an input image $I$, we transform it into a set of different views $V$,
including a global view $V_I$, and $N$ local views $\{ V_1, V_2, ..., V_N \}$,
which are randomly cropped from the global view.
The local views $\{ V_1, V_2, ..., V_N \}$ are sent into the local network 
focusing on generating attention maps that contain rich object details.
The global view $V_I$ is fed into the global network, which aims to
learn the knowledge from the local network and produces object attentions
in inference.
Let $\{ F_1, F_2, ..., F_N \}$ be the outputs of the last convolutional layer 
of the local network and each has $C$ channels
corresponding to the number of classes.
Let $\hat{F}$ be the output of the last convolutional layer of the global network 
that has $C+1$ channels.
The classification loss and the attention transfer loss can be defined
as follows.

\myPara{Classification Loss}
The classification loss is equipped with on the local network.
Specifically, the feature maps $\{ F_1, F_2, ..., F_N \}$ of the local views  
are first sent to a global pooling layer, 
where the features are pooled to a set of 1D feature vectors 
$\{ f_1, f_2, ..., f_N \}$.
Given a 1D feature vector $f_i$, 
the predicted probabilities for all categories can be 
computed by $q_i = \sigma(f_i)$.
Recall that $\sigma$ is the sigmoid function.
Then, the classification loss $L_{\text{cls}}$ can be written as
\begin{equation}
    L_{\text{cls}} = - \frac{1}{N\times C} \sum_{i=1}^N \sum_{c=1}^C y^c \log(q_i^c) + (1-y^c) \log(1 - q_i^c).   
\end{equation}

\myPara{Attention Transfer Loss}
We first generate attention maps for the local views from the local network. 
We use \eqnref{eqn:cam} to generate attention maps 
$\{ A_1^c, A_2^c, ..., A_N^c \}$ for the $c^{th}$ category
if $c$ is in the image-level labels.
If $c$ is not in the image-level labels, the attention values in the
corresponding attention map will be zeroed. 
To transfer the attentions attained by the local network to 
the global network, we adopt the mean squared error loss.

Given the output $\hat{F}$ from the global network, 
we apply a Softmax function to $\hat{F}$ along the channel dimension for
each location, yielding
\begin{equation}
    G^c = \frac{e^{\hat{F}^c}}{\sum_{i=1}^{C+1} e^{\hat{F}^i} },
\end{equation} 
where the value at each location of $G^c$ means the probability of 
this location being category $c$.
%
%
Let $\{G_1, G_2, ..., G_N \}$ denote the corresponding regions 
to $\{ A_1, A_2, ..., A_N \}$ on the global view, \ie, each pair
$(G_1, A_1)$ are cropped from the same coordinate on the global view.
The attention transfer loss is formulated by measuring the difference 
between $\{A_i\}$ and $\{G_i\}$ as follows:
\begin{equation}
    L_{\text{at}} = \frac{1}{N} \sum_{i=1}^N || A_i - G_i ||^2 .
\end{equation}
%
%
During training, we jointly optimize the above two losses.
During inference, the attention maps are generated from the 
global network while the local network can be discarded.
%
%

\myPara{Discussion} Our method provides an efficient way to
leverage the complementary information from the global view and
the local views.
%
%
%
The local-to-global attention transfer method conducts 
the global network to absorb the rich object detail knowledge 
captured by the local network in an online learning manner.
Though most previous works also use data augmentations, like random crop,
for the inputs, they do not have a component to
accumulate the object detail knowledge from the cropped local patches
online from a global view.
This makes our local-to-global strategy quite different from previous works.
We will show more advantages of the proposed approach over other methods
in the experiment section.


\newcommand{\addTexB}[1]{\contour{white}{\textcolor{black}{#1}}}
\newcommand{\addFigT}[2]{\begin{overpic}[width=0.120\textwidth,height=0.095\textwidth, ]{figure/attention/#1}\put(0.3,4){\addTexB{#2}}\end{overpic}}
\newcommand{\addFigsT}[2]{\addFigT{#1.jpg}{#2} & \addFigT{#1_att1.jpg}{} & \addFigT{#1_att2.jpg}{} & \addFigT{#1_att3.jpg}{} }

\begin{figure*}[t]
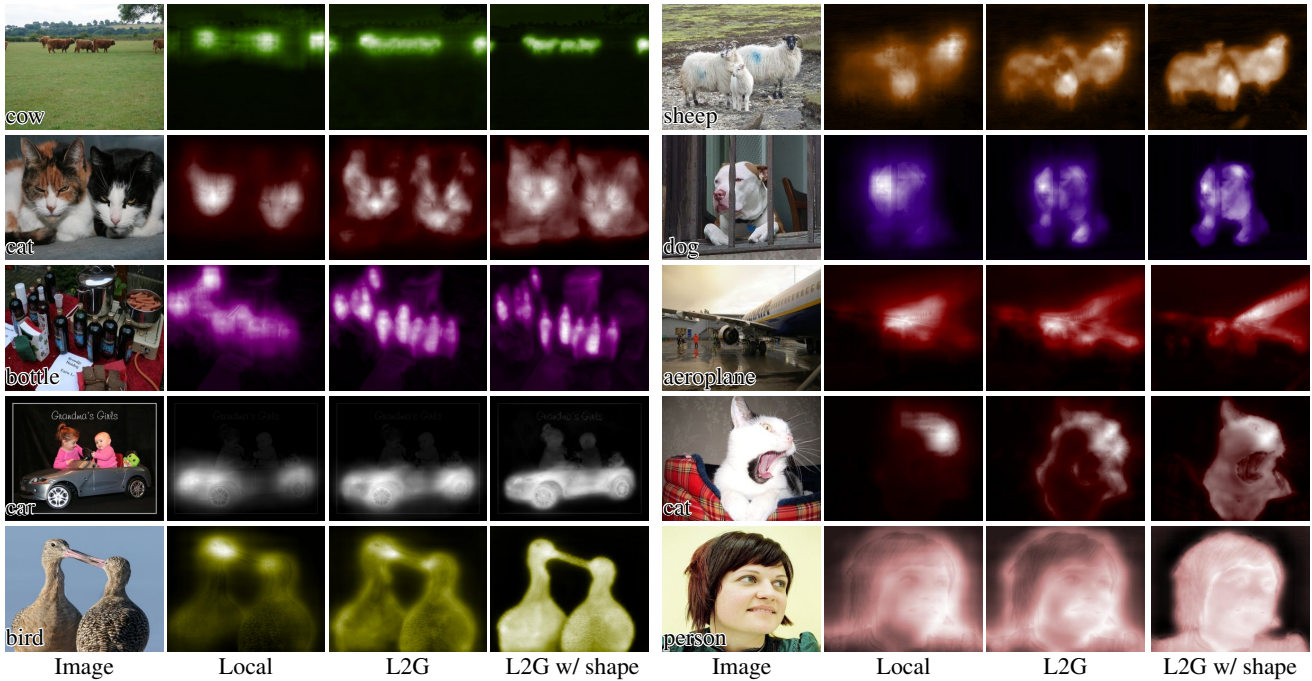
 
  \centering
  \small
  \setlength\tabcolsep{0.8pt}
  \renewcommand{\arraystretch}{0.7}
  \begin{tabular}{ccccccccc}
    \addFigsT{2007_001764}{cow} & & \addFigsT{2007_004423}{sheep} \\
    \addFigsT{2007_002760}{cat} & & \addFigsT{2007_001225}{dog} \\
    \addFigsT{2007_003207}{bottle} & &  \addFigsT{2007_004459}{aeroplane}\\
    \addFigsT{2007_004481}{car} & & \addFigsT{2007_003525}{cat}\\
    \addFigsT{2007_005264}{bird} & & \addFigsT{2007_005144}{person}\\
    Image & Local  & L2G & L2G w/ shape & & Image & Local  & L2G & L2G w/ shape \\
  \end{tabular}
  \vspace{-5pt}
  \caption{Qualitative comparison of attention maps from different methods.
  }
  \label{fig:cam}
  \vspace{-5pt}
\end{figure*}

\subsection{Local-to-Global Shape Transfer} \label{sec:st}
The proposed local-to-global attention transfer strategy can already
result in more integral object attentions than the original
CAM \cite{zhou2016learning}.
However, as the attention transfer process leverages only the image-level
labels, the captured attentions around the object boundaries 
are not sharp enough.
To well capture the shape of the localized objects in the attention maps, 
we attempt to introduce auxiliary salient object information
into the attention transfer loss by adding a shape constraint.
The saliency model \cite{liu2019simple} 
can serve as a class-agnostic salient object detector, 
which can segment the foreground objects and provide shape information.

The shape transfer process is simple, which has been illustrated in
the right part of \figref{fig:pipeline}.
%
%
Given the attention maps $\{ A_i \}$ from the local network,
we first binarize them with a small threshold (e.g., 0.1), yielding the binary maps $\{ B_i \}$.
Then, we utilize the saliency model to generate the saliency map $S$
for the given image $I$ and get the corresponding saliency regions to 
the attention maps $\{ A_i \}$ with the same coordinate on $I$
denoted as $\{ S_i \}$.
%
%
The attention transfer loss can then be rewritten as
\begin{equation} \label{eqn:st}
    L_{\text{st}} = \frac{1}{N} \sum_{i=1}^N  \begin{cases} 
        || B_i \times S_i - G_i ||^2, ~~\mbox{if} ~~|S_i| \neq 0
      \\  || A_i - G_i ||^2, ~~~~~~~~~~\mbox{if}~~|S_i| = 0
    \end{cases}
\end{equation}
where $\times$ denotes element-wise multiplication and 
$|S_i|$ is the cardinality of the saliency map $S_i$.
By using $B_i \times S_i$, we aim to remove the attention regions outside
the salient objects, which belong to the background with high probability. 
This allows our method to fully leverage the shape information provided
by the saliency maps and results in high-quality attention maps.
We will elaborate more on this in our experiment section.
Note that as not all the images would have salient objects,
it is inappropriate to always use the top part of \eqnref{eqn:st}.
Thus, for those images whose saliency maps contain nothing,
we utilize the original attention maps as supervision as formulated in
the bottom part of \eqnref{eqn:st}.

It is worth mentioning that EPS~\cite{lee2021railroad} also uses
saliency maps as supervision to provide the network with shape
information.
Differently, our method focuses more on how to take advantage of
multiple local views and how to efficiently transfer the learned knowledge
from the local network to the global one.
In the following, we will show the advantage of the
proposed local-to-global knowledge transfer over EPS.

\section{Experiments} 
The following paper is organized as follows.
First, we introduce the experimental setup.
Then, we show experimental results on ablation study
and analyze the role of each component proposed in our method.
Finally, we conduct experiments to compare our method with 
previous state-of-the-art WSSS methods.

\subsection{Experimental Setup} \label{sec:exp_setup}
\myParaP{Datasets.}
Experiments are conducted on two publicly available datasets, PASCAL VOC 2012
and MS COCO 2014.
The PASCAL VOC 2012 dataset contains 20 semantic categories and the background.
It is split into three sets, the training, validation, and test sets, 
each containing 1464, 1449, and 1456 images, respectively.
Following most previous works, we also use the augmented training set 
\cite{hariharan2011semantic}, yielding totally 10582 training images.
The MS COCO 2014 dataset has 80 semantic categories. 
Following \cite{choe2020attention,lee2021railroad}, the images without 
target categories are excluded from the dataset, remaining 
82081 training images and 40137 validation images. 

\myParaP{Evaluation metric.}
The mean intersection-over-union (mIoU) \cite{long2015fully} 
is used as the evaluation metric.
As the segmentation annotations of the test set in the PASCAL VOC 2012 dataset
are not available, we submit the segmentation results to the official PASCAL VOC 
evaluation server\footnote{http://host.robots.ox.ac.uk:8080/}.

\myParaP{Data augmentation.}
For data augmentation, the short size of the input image 
is resized to 512.
The global view is with a resolution of 448$\times$448, which
is cropped from the input image.
The local image patches with resolution 320$\times$320 are 
cropped from the global view.

\myParaP{Classification Network.}
Following \cite{ahn2018learning,lee2021railroad}, we utilize 
ResNet-38 \cite{wu2019wider} as our classification network.
Besides, we also employ a pixel correlation module (PCM) \cite{wang2020self} 
into the classification network to constrain the shape of 
the target object.
The attention maps are generated from the global network 
using the multi-scale test strategy~\cite{ahn2018learning}.

\myParaP{Classification on PASCAL VOC.}
We train the classification network for 10 epochs 
and use SGD as the optimizer.
The initial learning rate is set to 1e-3, 
which decays at the $6^{th}$ epoch.
The loss weight $\lambda$ for the attention transfer 
loss is 10.
Other network settings are as follows:
batch size: 3, weight decay: 5e-4, patch 
size: 320$\times$320, patch number: 6.

\myParaP{Classification on MS COCO.}
We train the classification network for 15 epochs 
and use SGD as the optimizer.
We set the initial learning rate to 0.1 
and use \emph{poly} as the learning scheduler.
The loss weight $\lambda$ for the attention transfer 
loss is 30.
Other network settings are as follows:
batch size: 12, weight decay: 5e-4, 
patch size: 320$\times$320, patch number: 4.

\myParaP{Segmentation.}
We select DeepLab-v1~\cite{chen2014semantic} 
and DeepLab-v2 \cite{chen2017deeplab} 
as our segmentation networks.
We report performance based on both VGG-16 \cite{simonyan2014very} 
and ResNet-101 \cite{he2016deep}.
For VGG-16 based segmentation network, we use the classification 
model pretrained on ImageNet \cite{deng2009large} 
for initialization.
For ResNet-101, we use the COCO pretrained model. 
For the experiments on MS COCO dataset, we all 
utilize ImageNet pretrained model.
%
Following \cite{lee2021railroad}, we use the 
same way to generate 
the pseudo labels.
Given the attention maps, we assign a 
fixed threshold to the background channel 
and use the argmax function to obtain the label 
for each pixel.

\begin{table}[t]
  \centering
  \small
  \renewcommand{\arraystretch}{1.1}
  \setlength\tabcolsep{2.35mm}
  \caption{Comparisons of mIoU scores under different network settings.
  The baseline is the original CAM~\cite{zhou2016learning}.
  SW: The sliding window strategy is applied to the baseline
  during inference~\cite{zhou2016learning}.
  Local: Using multiple local image patches instead of the 
  input image to train the classification network.
  L2G: Our method with local-to-global attention transfer only.
  $\mbox{mIoU}_{trainaug}$ denotes the mIoU score of the pseudo segmentation labels on the augmented training set.
  } 
  \vspace{-5pt}
  \begin{tabular}{c|ccc|cc} \toprule[1.0pt]
    No.    & SW           &   Local & L2G       & $\mbox{mIoU}_{trainaug}$  & $\mbox{mIoU}_{val}$   \\ 
    \midrule[0.8pt]
    1     &               &            &     &    47.1         &   47.5    \\ 
    2     &  \checkmark   &            &     &    46.1 (\highlight{-1.0})        &   47.2 (\highlight{-0.3})        \\ 
    3     &               & \checkmark &     &    48.5 (\highlight{+1.4})        &   50.0 (\highlight{+2.5})   \\ 
    4     &               &  &  \checkmark   &    56.8 (\highlight{+9.7})        &   54.9 (\highlight{+7.4})   \\ 
 \bottomrule[1.0pt]
  \end{tabular}
  \vspace{-10pt}
  \label{tab:abla1}
\end{table}

\begin{table}[t]
  \centering
  \small
  \caption{Ablation experiments on the importance of each
  component. L2G: Local-to-global attention transfer only.
  Shape: Local-to-global shape transfer. As can be seen,
  our local-to-global transfer strategy can significantly improve
  the performance compared to the setting with only the local network
  being used. When the shape information is incorporated, L2G
  can still lift the performance by a large margin.
  } 
  \vspace{-5pt}
  \renewcommand{\arraystretch}{1.1}
  \setlength\tabcolsep{2mm}
  \begin{tabular}{c|ccc|cc} \toprule[1.0pt]
    No.   &   Local       &   L2G          &   Shape      & $\mbox{mIoU}_{trainaug}$  & $\mbox{mIoU}_{val}$   \\ 
    \midrule[0.8pt]
    1     & \checkmark   &                &              &    48.5         &   50.0    \\ 
    2     &              & \checkmark     &              &    56.8 (\highlight{+8.3})         &   54.9 (\highlight{+4.9})   \\ \midrule[0.5pt]
    3     & \checkmark   &                & \checkmark   &    68.0         &   69.9   \\
    4     &              & \checkmark     & \checkmark   &    70.3 (\highlight{+2.3})        &   72.1 (\highlight{+2.2})    \\
 \bottomrule[1.0pt]
  \end{tabular}
  \vspace{-10pt}
  
  \label{tab:abla2}
\end{table}

\subsection{Ablation Study}  \label{sec:exp_abla}
We design multiple ablation experiments 
to perform a sanity check for our method.
All the ablation experiments are conducted on the
PASCAL VOC 2012 dataset. 
We report the mIoU scores of the pseudo segmentation labels 
on the augmented training set and the mIoU scores of 
the segmentation results on the validation set.

\myParaP{Local view sampling strategy.}
First, we study the impact of the sampling strategy on the attention maps.
We compare two local view sampling strategies. 
One is the random sampling strategy, and the other is the uniform sampling strategy.
We implement the uniform sampling strategy by sliding the window over
the global view uniformly.
In this way, every pixel can be enclosed within some local view.
For the global view with 448$\times$448 resolution, 
we set the window size to 320$\times$320 and the stride to 64, obtaining 9 local views.
For a fair comparison with the uniform sampling strategy, 
we randomly sample 9 image patches for the random sampling strategy.
The qualities of the pseudo segmentation labels using these two strategies 
are quite close to each other (random 68.8\% v.s. uniform 68.5\%).
To flexibly adjust the local view number $N$, we choose 
the random sampling strategy in our method.

\begin{figure}[t]
  \centering
  \setlength\tabcolsep{1pt}
  \includegraphics[width=0.235\textwidth]{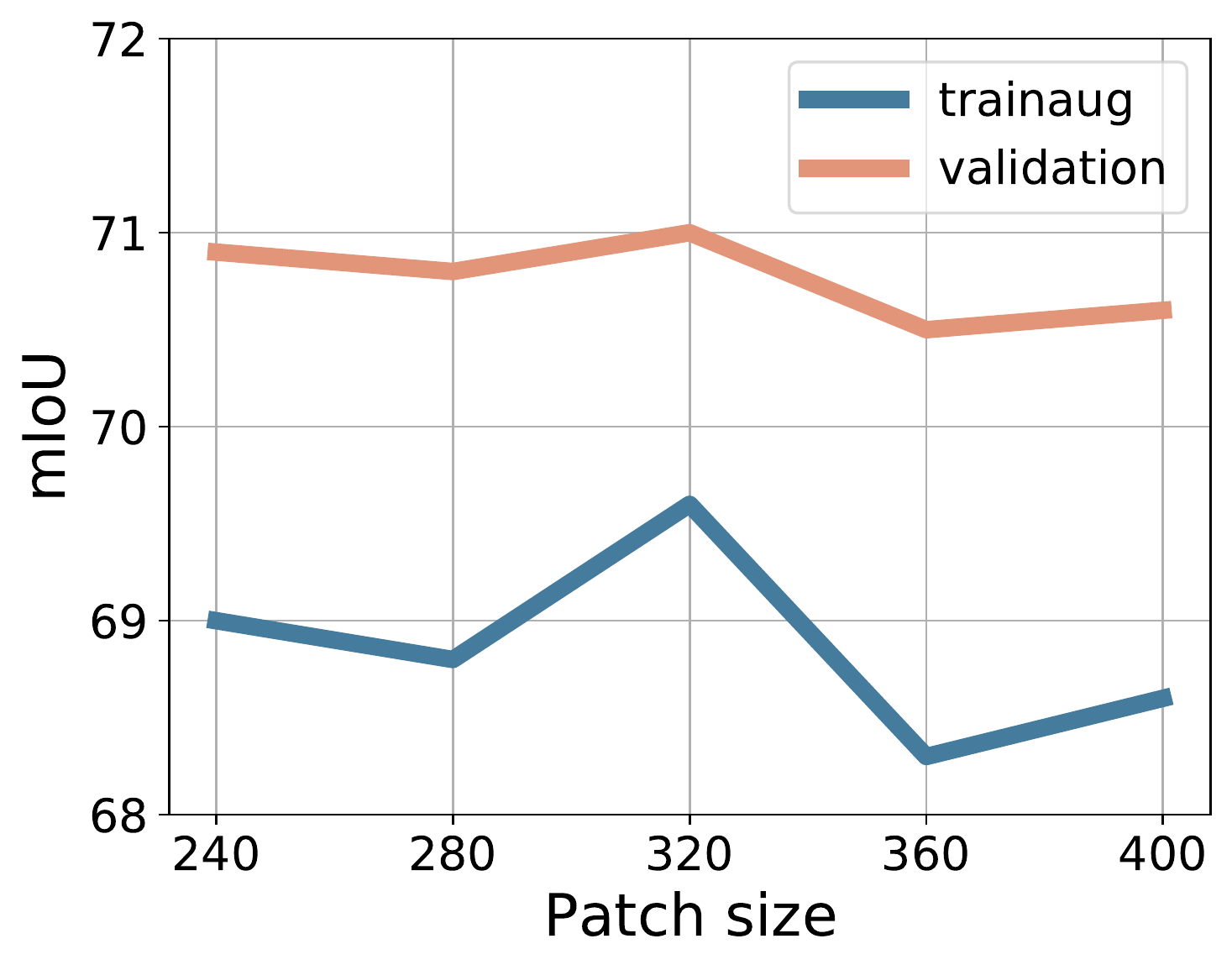}
  \includegraphics[width=0.235\textwidth]{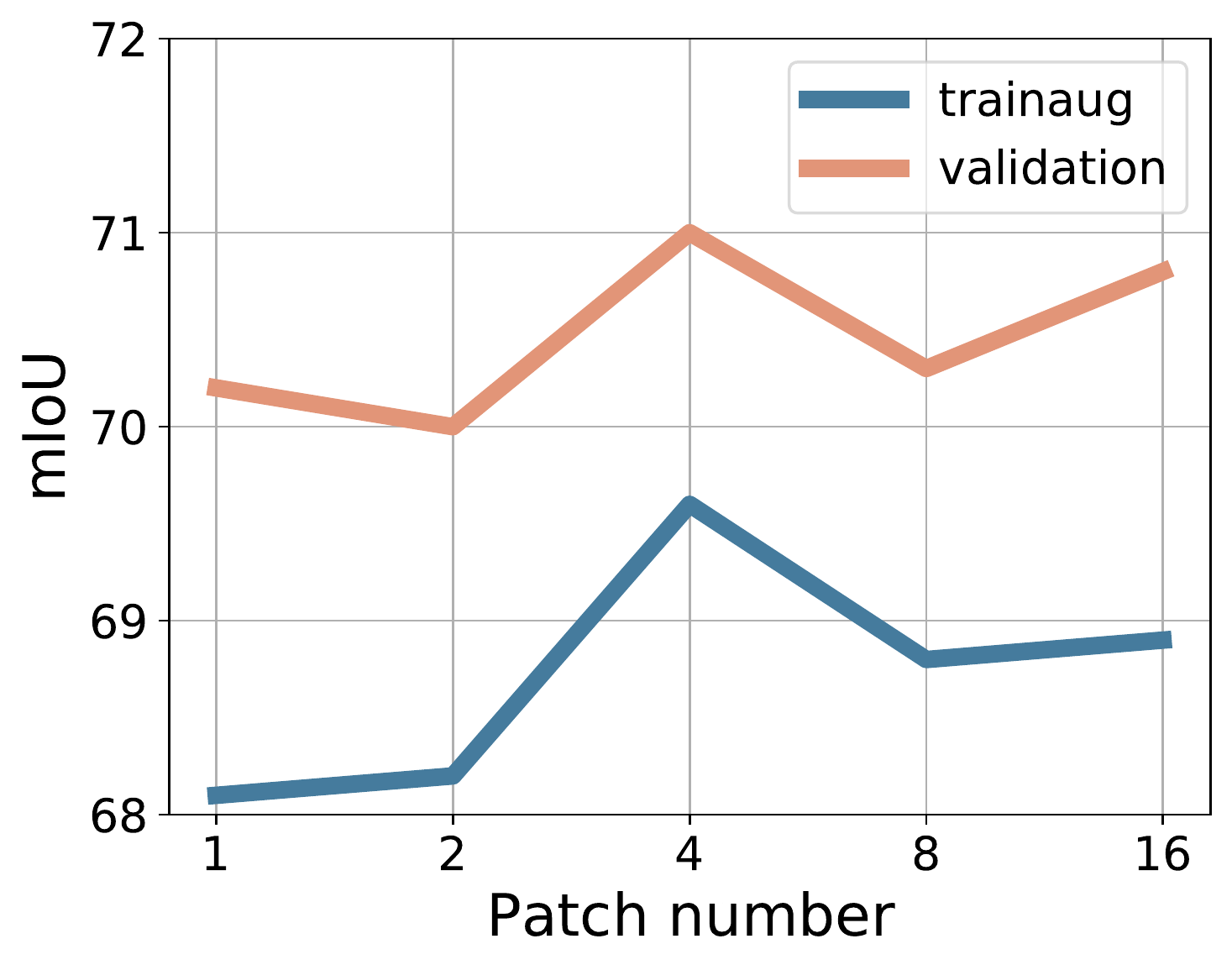}
  \vspace{-15pt}
  \caption{Ablations on the local view size and number $N$.
  }\label{fig:abla3}
  \vspace{-5pt}
\end{figure}

\myParaP{Patch size and patch number $N$. } 
The patch size controls the spatial size of the local views.
The patch number $N$ denotes the number of the local views sent to the local network.
%
%
To study their impact on the attention quality,
we select 5 different patch sizes 
[240$\times$240, 280$\times$280, 320$\times$320, 360$\times$360, 400$\times$400].
When studying the patch number $N$, we select the number of 
local views from the range of [1, 2, 4, 8, 16].
%
%
As shown in \figref{fig:abla3}, 
we observe that when $N$ increases, the quality of 
the pseudo segmentation labels becomes better. 
The performance tends to be robust when the local view number
is larger than 4.
For the patch size, we can see that our method 
achieves the best performance when the local view size
takes 320$\times$320.
When the size is larger than 320$\times$320, the quality of 
the pseudo segmentation labels decreases largely.

\renewcommand{\addFig}[1]{\includegraphics[width=0.14\linewidth]{figure/seg/#1}}
\renewcommand{\addFigs}[1]{\addFig{#1.jpg}  & \addFig{#1_1.png} & \addFig{#1_2.png}   & \addFig{#1_3.png} 
&  \addFig{#1_4.png} & \addFig{#1_5.png} & \addFig{#1.png} }

\begin{figure*}[t] 
  \centering
  \small
  \setlength\tabcolsep{0.6pt}
  \renewcommand{\arraystretch}{0.7}
  \begin{tabular}{ccccccc}
    \addFigs{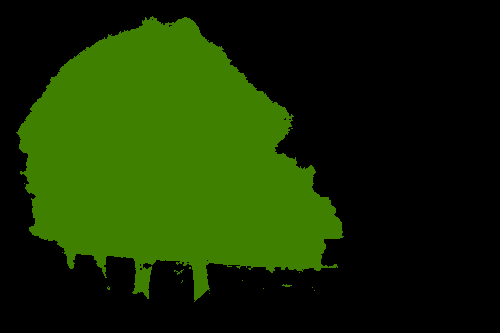} \\
    \addFigs{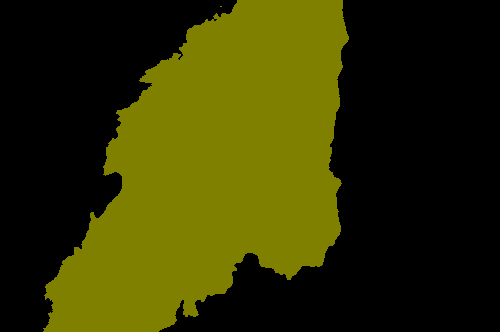} \\
    \addFigs{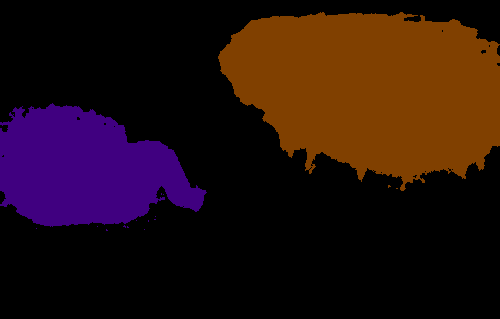} \\
    \addFigs{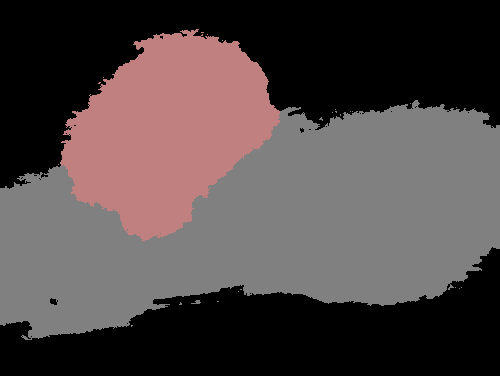} \\
    \addFigs{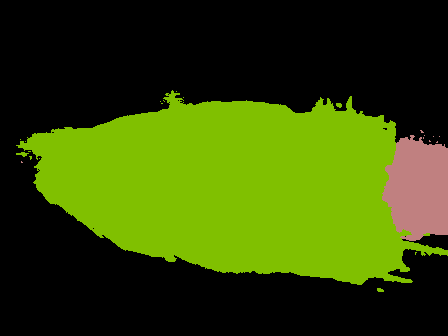} \\
    \addFigs{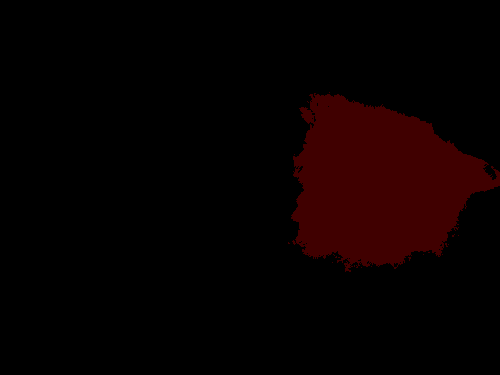} \\
    \addFigs{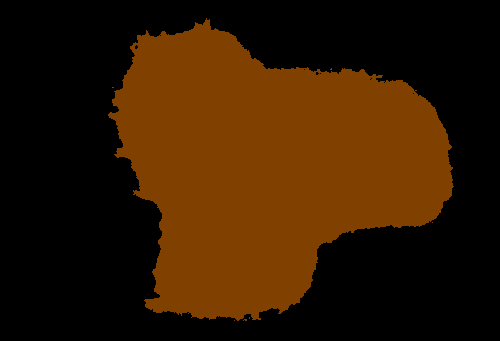} \\
    \addFigs{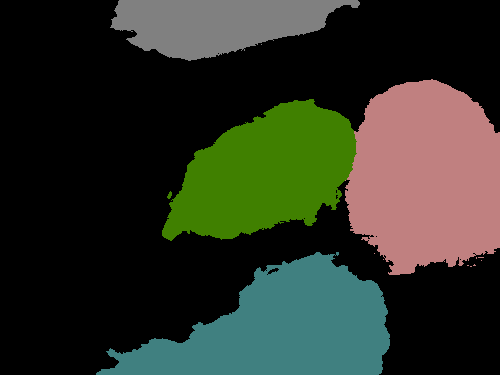} \\
    Image & CAM   & Local &  L2G  & Local + Shape & L2G + Shape & GT \\
  \end{tabular}
  \vspace{-5pt}
  \caption{Comparison of the segmentation results under different network settings. We can observe that the combining L2G and
  shape transfer yields the best results, especially on local object details. }
  \vspace{-5pt}
  \label{fig:seg}
\end{figure*}

\myParaP{Importance of the proposed local-to-global knowledge transfer. }
When sending the local views to the local network, 
we can discover more object regions
from the resulting attention maps.
%
%
Here, one may raise a question: ``Are the attention maps 
from the local network good enough so that we do not need the transfer process?''.
To answer this question, we test the quality of 
the pseudo segmentation labels using the attention maps 
from the local network.
As shown in \tabref{tab:abla1}, we can see the performance of 
the local network is slightly better than the baseline CAM \cite{zhou2016learning}.
However, the performance is far lower than L2G (48.5\% v.s. 56.8\%).
We also show some qualitative  results of the attention maps 
in \figref{fig:cam} and segmentation results in \figref{fig:seg}.
This indicates the local-to-global attention transfer strategy 
is a more efficient way to leverage the rich object attention
knowledge captured by the local network.

In addition, we further extend the above experiments by
introducing the shape information.
The corresponding results can be found in \tabref{tab:abla2}.
It has been demonstrated in \cite{lee2021railroad} that
saliency shape information can significantly improve attention
quality.
However, when the proposed L2G is used, the mIoU scores on
both the trainaug and validation sets  can be
largely improved.
We will show more numerical results on segmentation in the next
subsection.

\myParaP{L2G v.s. Sliding window. } 
%
%
The key of our method is to leverage the local attention maps 
to facilitate the global network to discover more integral
object regions.
One direct way to implement this idea is to utilize 
the sliding window strategy during inference and aggregate the
attention maps from different image patches.
We compare our L2G with the sliding window strategy.
Specifically, for the sliding window strategy, the window size 
and the stride are set to 320$\times$320 and 64, respectively.
For our L2G, we set the local view size 
to 320$\times$320 and the number of randomly sampled patches
to 4 at a time.

As shown in \tabref{tab:abla1}, the local-to-global attention transfer strategy 
achieves much better results than the baseline CAM \cite{zhou2016learning}, 
which verifies the effectiveness of our method.
However, the results of the sliding window strategy are even
worse than the original CAM.
We argue that the sliding window strategy is not suitable 
for mining non-discriminative object regions as the trained model
is still based on inputs with the global view.
This makes the undiscovered object regions
that have different appearances with the distinctive 
regions hard to respond when processing the global view.
%
%

\myParaP{Classification loss in the global network. } 
The local network is equipped with the 
classification loss to guide the attention generation.
One may ask the question ``Does the global network also need the classification loss?''
To answer this question, we have attempted to add
a classification loss to the global network.
%
%
We observe that the attention maps generated from the global network 
locate very small object regions when adding the classification loss.
The quality of the pseudo segmentation labels decreases largely from 70.3\% to 53.8\%.
We argue that the classification loss and the attention transfer loss play opposite roles.
The classification loss makes the attention be more discriminative.
The attention transfer loss aids in transferring the attention 
on non-discriminative regions to the global network.
Thus, the attention maps become worse.

\myParaP{Local and global backbone sharing. } 
Here, we explore the performance gap between with/without 
the local and global network backbone sharing.
When the local and global networks share the same backbone, 
the mIoU score of the pseudo segmentation labels on the trainaug set is 69.2\%.
After training the segmentation network, the mIoU score on the validation set is 70.9\%.
When the local and global networks utilize different backbones, 
the mIoU score of the pseudo segmentation labels can be improved by 1.1\%.
The final segmentation result also attains 1.2\% mIoU gains.

\begin{table}[t]
  \small
  ~~~\begin{minipage}[c]{0.2\textwidth}
  \centering
  \caption{Comparison of pseudo segmentation labels 
  on the PASCAL VOC train set with no saliency maps. }
  \label{tab:seed_comps}
  \vspace{-5pt}
  \renewcommand{\arraystretch}{1.1}
  \setlength\tabcolsep{1.5mm}
  \begin{tabular}{l|c} \toprule[1.0pt]
     Methods     & $\mbox{mIoU}_{train}$    \\ 
     \midrule[0.8pt]
     CAM \cite{zhou2016learning}     &  48.0  \\  
     SC-CAM \cite{chang2020weakly}   &  50.9  \\  
     SEAM \cite{wang2020self}        &  55.4  \\
     ADvCAM \cite{lee2021anti}       &  55.6  \\
     \rowcolor{gray!32} L2G (ours)   &  56.2  \\
  \bottomrule[1.0pt]
  \end{tabular}    
  \vspace{-10pt} 
  \end{minipage}  ~~~~~~~~~
  \begin{minipage}[c]{0.2\textwidth}
  \centering
  \caption{Comparisons of pseudo segmentation labels 
  on the PASCAL VOC train set with saliency maps incorporated.
  }
  \renewcommand{\arraystretch}{1.3}
  \setlength\tabcolsep{1.5mm}
  \begin{tabular}{l|c} \toprule[1.0pt]
   Methods     & $\mbox{mIoU}_{train}$    \\ 
   \midrule[0.8pt]
   SGAN \cite{yao2020saliency}    &  62.8  \\
   EPS \cite{lee2021railroad}     &  69.4  \\
   \rowcolor{gray!32} L2G (ours)  &  71.9  \\
  \bottomrule[1pt]
  \end{tabular}   
  \vspace{-10pt} 
\label{tab:seed_comps_sal}
\end{minipage} 
\end{table}

\begin{table}[t]
    \centering
    \small
        \caption{Quantitative comparisons to previous state-of-the-art approaches
        on PASCAL VOC 2012 validation and test sets. 
        All the segmentation results are based on the DeepLab with VGGNet backbone \cite{simonyan2014very}. 
        Pub.: Publication,
        Seg.:Segmentation network,
        Sup.: Supervision, I.: Image-level label, 
        S.: saliency maps from the off-the-shelf saliency model.
        }\label{tab:comps_pascal_vgg}
        \vspace{-5pt}
    \renewcommand{\arraystretch}{1.10}
    \setlength{\tabcolsep}{1.0mm}{
    \begin{tabular}{l|c|c|c|c|c} \toprule[1pt] 
        Methods                                    & Pub.     & Seg.    & Sup.   & Val (\%)   & Test (\%) 
        \\ \midrule[0.8pt]
        AffinityNet \cite{ahn2018learning}         &  CVPR'18 & V1 & I.     & 58.4       & 60.5 \\
        MCOF \cite{wang2018weakly}                 &  CVPR'18 & V1   & I.+S.  & 56.2       & 57.6 \\ 
        DSRG \cite{huang2018weakly}                &  CVPR'18 & V2   & I.+S.  & 59.0       & 60.4 \\
        SeeNet \cite{hou2018self}                  &  NeurIPS'18&V1  & I.+S.  & 61.1       & 60.7 \\
        FickleNet \cite{lee2019ficklenet}          &  CVPR'19 & V2   & I.+S.  & 61.2       & 61.9 \\
        $\mbox{OAA}^{+}$ \cite{jiang2019integral}  &  ICCV'19 & V1   & I.+S.  & 63.1       & 62.8 \\
        BES \cite{chen2020weakly}                  &  ECCV'20 & V1   & I.     & 60.1       & 61.1 \\
        MCIS \cite{sun2020mining}                  &  ECCV'20 & V1   & I.+S.  & 63.5       & 63.6 \\
        Multi-Est. \cite{fan2020employing}         &  ECCV'20 & V1   & I.+S.  & 64.6       & 64.2 \\
        ICD \cite{fan2020learning}                 &  CVPR'20 & V1   & I.+S.  & 64.0       & 63.9 \\
        ECS-Net \cite{sun2021ecs}                  &  ICCV'21 & V1   & I.     & 62.1       & 63.4 \\
        DRS \cite{kim2021discriminative}           &  AAAI'21 & V1   & I.+S.  & 63.5       & 64.5 \\
        Group-WSSS \cite{li2021group}              &  AAAI'21 & V2   & I.+S.  & 63.3       & 63.6 \\
        $\mbox{OAA++}^{+}$ \cite{jiang2021online}  &  PAMI'21 & V1  & I.+S.  & 63.7       & 63.2 \\
        NSROM \cite{yao2021non}                    &  CVPR'21 & V2   & I.+S.  & 65.5       & 65.3 \\
        EPS \cite{lee2021railroad}                 &  CVPR'21 & V1   & I.+S.  & 66.6       & 67.9 \\
        EPS \cite{lee2021railroad}                 &  CVPR'21 & V2   & I.+S.  & 67.0       & 67.3 \\
        \rowcolor{gray!32} L2G (ours)              &  --      & V1   & I.+S.  & 68.1       &  68.8    \\
        \rowcolor{gray!32} L2G (ours)              &  --      & V2   & I.+S.  & 68.5       &  68.9    \\
        \bottomrule[1pt]
        \end{tabular} }
        \vspace{-10pt}
\end{table}

\subsection{Comparisons with the State-of-the-Arts} \label{sec:exp_comp}
We first compare the quality of our produced attention maps
with the previous state-of-the-art WSSS methods.
Our attention maps are converted to pseudo segmentation labels.
%
As shown in \tabref{tab:seed_comps} and \tabref{tab:seed_comps_sal}, 
it is obvious that the attention maps generated by our method are 
better than other methods no matter whether the saliency maps
are used.
Without the saliency maps, the mIoU score on the PASCAL VOC train set reaches 56.2\%, better than SEAM \cite{wang2020self} by 0.8\%. 
After applying the saliency maps to the transfer process, 
the mIoU score reaches 71.9\%, much better than
EPS \cite{lee2021railroad} (71.9\% v.s. 69.4\%).

We use the pseudo segmentation labels to train the DeepLab 
segmentation model directly.
We compare the segmentation performance of our method with previous 
state-of-the-art methods.
\tabref{tab:comps_pascal_vgg} and \tabref{tab:comps_pascal_resnet} list 
the segmentation results of our method and the recent state-of-the-art methods 
on the PASCAL VOC dataset.
As we can see, compared to the previous WSSS methods, 
our method achieves the best results on both the validation and
test sets.
The work most relevant to our method is EPS \cite{lee2021railroad}, 
which explicitly uses the saliency maps as supervision.
The differences between our method and EPS have been explained in
\secref{sec:st}.
%
%
%
As shown in \tabref{tab:comps_pascal_resnet}, 
we can see that our method can improve 
the results of EPS by around 1\%.
Besides, as shown in \tabref{tab:comps_coco}, our results on the 
challenging MS COCO dataset are much better than the previous methods,
which also demonstrates the effectiveness 
of our local-to-global strategy.
The mIoU of the pseudo labels for our method is 43.4\%, 
much better than that of EPS (37.2\%).

\begin{table}[t]
    \centering
    \small
        \caption{Quantitative comparisons to previous state-of-the-art approaches
        on PASCAL VOC 2012 validation and test sets. 
        All the segmentation results are based on the
        ResNet backbone \cite{he2016deep,wu2019wider}.
        Our method achieves the best results.
        }\label{tab:comps_pascal_resnet}
        \vspace{-5pt}
    \renewcommand{\arraystretch}{1.10}
    \setlength{\tabcolsep}{1.mm}{
    \begin{tabular}{l|c|c|c|c|c} \toprule[1pt] 
        Methods                                     &   Publication     & Seg. & Sup.   & Val (\%)   & Test (\%) 
        \\ \midrule[0.8pt]
        AffinityNet \cite{ahn2018learning}          & CVPR'18    & V1  & I.     & 61.7       & 63.7 \\
        MCOF \cite{wang2018weakly}                  & CVPR'18    & V1  & I.+S.  & 60.3       & 61.2 \\ 
        DSRG \cite{huang2018weakly}                 & CVPR'18    & V2  & I.+S.  & 61.4       & 63.2 \\ 
        SeeNet \cite{hou2018self}                   & NeurIPS'18 & V1  & I.+S.  & 63.1       & 62.8 \\
        IRNet \cite{ahn2019weakly}                  & CVPR'19    & V1  & I.     & 63.5       & 64.8 \\
        FickleNet \cite{lee2019ficklenet}           & CVPR'19    & V2  & I.+S.  & 64.9       & 65.3 \\
        $\mbox{OAA}^{+}$ \cite{jiang2019integral}   & ICCV'19    & V1  & I.+S.  & 65.2       & 66.4 \\
        SSDD \cite{shimoda2019self}                 & ICCV'19    & V1  & I.     & 66.1       & 66.8 \\ 
        SEAM \cite{wang2020self}                    & CVPR'20    & V2  & I.     & 64.5       & 65.7 \\
        SC-CAM \cite{chang2020weakly}               & CVPR'20    & V2  & I.     & 66.1       & 65.9 \\
        ICD \cite{fan2020learning}                  & CVPR'20    & V1  & I.+S.  & 67.8       & 68.0 \\
        BES \cite{chen2020weakly}                   & ECCV'20    & V2  & I.     & 65.7       & 66.6 \\
        MCIS \cite{sun2020mining}                   & ECCV'20    & V1  & I.+S.  & 66.2       & 66.9 \\
        Multi-Est. \cite{fan2020employing}          & ECCV'20    & V1  & I.+S.  & 67.2       & 66.7 \\
        LIID \cite{liu2020leveraging}               & PAMI'20    & V2  & I.+IS. & 66.5       & 67.5 \\
        DRS \cite{kim2021discriminative}            & AAAI'21    & V2  & I.+S.  & 71.2       & 71.4 \\
        Group-WSSS \cite{li2021group}               & AAAI'21    & V2  & I.+S.  & 68.2       & 68.5 \\
        ECS-Net \cite{sun2021ecs}                   & ICCV'21    & V1  & I.     & 66.6       & 67.6 \\  
        PMM \cite{li2021pseudo}                     & ICCV'21    & PSP & I.     & 68.5       & 69.0 \\
        CDA \cite{su2021context}                    & ICCV'21    & V2  & I.     & 66.1       & 66.8 \\
        CGNet \cite{kweon2021unlocking}             & ICCV'21    & V1  & I.     & 68.4       & 68.2 \\  
        AuxSegNet \cite{xu2021leveraging}           & ICCV'21    & V1  & I.+S.  & 69.0       & 68.6 \\
        AdvCAM \cite{lee2021anti}                   & CVPR'21    & V2  & I.     & 68.1       & 68.0 \\       
        NSROM \cite{yao2021non}                     & CVPR'21    & V2  & I.+S.  & 70.4       & 70.2 \\ 
        EDAM \cite{wu2021embedded}                  & CVPR'21    & V1  & I.+S.  & 70.9       & 70.6 \\
        EPS \cite{lee2021railroad}                  & CVPR'21    & V1  & I.+S.  & 71.0       & 71.8 \\
        EPS \cite{lee2021railroad}                  & CVPR'21    & V2  & I.+S.  & 70.9       & 70.8 \\
        \rowcolor{gray!32} L2G (ours)               &  --        & V1 & I.+S.  & 72.0       & 73.0 \\
        \rowcolor{gray!32} L2G (ours)               &  --        & V2 & I.+S.  & 72.1       & 71.7 \\
        \bottomrule[1pt]
        \end{tabular} }
        \vspace{-5pt}
\end{table}

\begin{table}[h]
    \centering
    \small
    \renewcommand{\arraystretch}{1.10}
    \caption{Quantitative comparisons to previous state-of-the-art approaches
    on MS COCO validation set. 
    All the segmentation results are based on VGGNet  backbone \cite{simonyan2014very} except L2G* using ResNet-101 backbone \cite{he2016deep}.
    }\label{tab:comps_coco}
    \vspace{-5pt}
    \setlength{\tabcolsep}{3.0mm}{
    \begin{tabular}{l|c|c|c|c} \toprule[1pt] 
        Methods                                      & Publication             & Seg. & Sup.   & Val (\%)    \\ \midrule[0.8pt]
        SEC \cite{kolesnikov2016seed}               & ECCV'16  &   V1   & I.+S.  & 22.4        \\
        DSRG \cite{huang2018weakly}                 & CVPR'18  &   V2   & I.+S.  & 26.0        \\ 
        ADL \cite{choe2020attention}                & PAMI'20  &   V1   & I.+S.  & 30.8        \\ 
        EPS \cite{lee2021railroad}                  & CVPR'21  &   V2   & I.+S.  & 35.7        \\
        \rowcolor{gray!32} L2G (ours)               & --       &   V2   & I.+S.  &  42.7           \\
        \rowcolor{gray!32} L2G* (ours)               & --      &   V2   & I.+S. &  44.2           \\
        \midrule[1pt]
        \end{tabular} }
        \vspace{-10pt}
\end{table}

\myParaP{Discussion.} It is worthy to note that our local network 
is just a simple classification model.
Because of the flexibility of the proposed framework, we can replace
the local network with more complicated attention models to further
improve the results.
Thus, we believe there is still a large room to improve the proposed
framework, and we also hope our local-to-global knowledge transfer
method could provide researchers with a new research direction.

\myParaP{Analysis of failure cases.} 
First, some non-target objects are wrongly recognized as
the target classes as shown in the first two rows of \figref{fig:failure}. 
In our L2G, we only use ResNet to extract attention. 
Designing more advanced classification models, such as 
transformers~\cite{dosovitskiy2020image,yuan2021tokens}, 
could improve the results.
Second, the shape of the discovered objects is still being
further improved (the last two rows). 
Using stronger saliency models or over-segmentation methods could, 
to some extent, solve this.

\renewcommand{\addFig}[1]{\includegraphics[width=0.32\linewidth]{figure/failure/#1}}
\renewcommand{\addFigs}[1]{\addFig{#1.jpg}   &  \addFig{#1_p.png}  & \addFig{#1.png} }

\begin{figure}[h] 
  \centering
  \small
  \setlength\tabcolsep{0.6pt}
  \renewcommand{\arraystretch}{0.8}
  \begin{tabular}{ccc}
    \addFigs{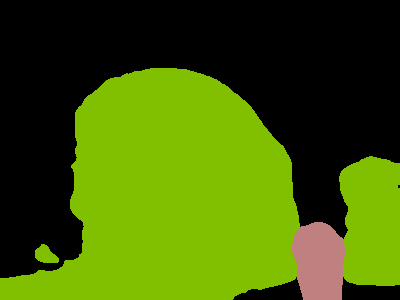} \\
    \addFigs{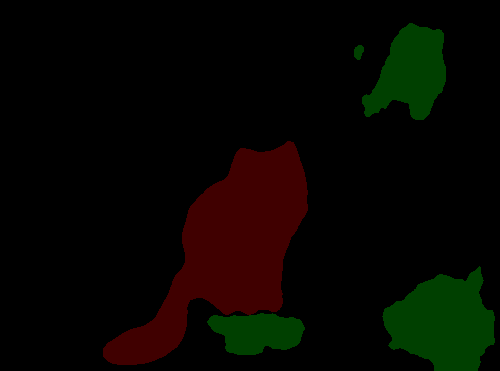} \\
    \addFigs{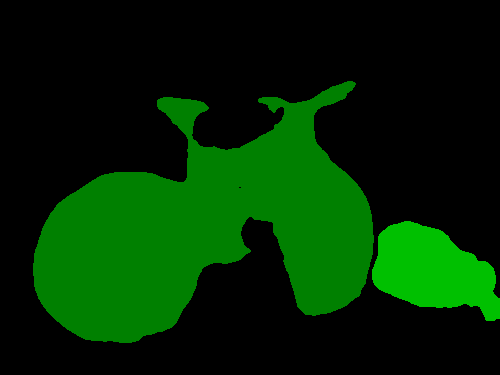} \\
    \addFigs{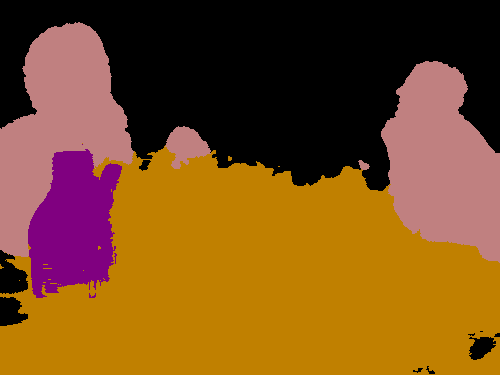} \\
    Image & Result  & GT   \\
  \end{tabular}
     \vspace{-5pt}
    \caption{Two failure segmentation examples of our L2G.
    }\label{fig:failure}
    \vspace{-15pt}
\end{figure}

\section{Conclusion}
In this paper, we propose a novel local-to-global attention transfer
method to attain object attentions.
By leveraging the complementary attention captured by the
local network from the local views and introducing the
shape constraint to the attention transfer process,
our method achieves the best results on both the validation
and test sets of PASCAL VOC 2012 and the validation set of 
MS COCO 2014.
We hope the proposed approach could facilitate the research
on vision tasks relying on high-quality attention maps.

{\small
\bibliographystyle{ieee_fullname}
\bibliography{l2g}

\begin{thebibliography}{10}\itemsep=-1pt

\bibitem{ahn2019weakly}
Jiwoon Ahn, Sunghyun Cho, and Suha Kwak.
\newblock Weakly supervised learning of instance segmentation with inter-pixel
  relations.
\newblock In {\em {IEEE Conf. Comput. Vis. Pattern Recog.}}, pages 2209--2218,
  2019.

\bibitem{ahn2018learning}
Jiwoon Ahn and Suha Kwak.
\newblock Learning pixel-level semantic affinity with image-level supervision
  for weakly supervised semantic segmentation.
\newblock In {\em {IEEE Conf. Comput. Vis. Pattern Recog.}}, pages 4981--4990,
  2018.

\bibitem{araslanov2020single}
Nikita Araslanov and Stefan Roth.
\newblock Single-stage semantic segmentation from image labels.
\newblock In {\em {IEEE Conf. Comput. Vis. Pattern Recog.}}, pages 4253--4262,
  2020.

\bibitem{bearman2016s}
Amy Bearman, Olga Russakovsky, Vittorio Ferrari, and Li Fei-Fei.
\newblock What’s the point: Semantic segmentation with point supervision.
\newblock In {\em {Eur. Conf. Comput. Vis.}}, pages 549--565, 2016.

\bibitem{chang2020mixup}
Yu-Ting Chang, Qiaosong Wang, Wei-Chih Hung, Robinson Piramuthu, Yi-Hsuan Tsai,
  and Ming-Hsuan Yang.
\newblock Mixup-cam: Weakly-supervised semantic segmentation via uncertainty
  regularization.
\newblock In {\em {Brit. Mach. Vis. Conf.}}, 2020.

\bibitem{chang2020weakly}
Yu-Ting Chang, Qiaosong Wang, Wei-Chih Hung, Robinson Piramuthu, Yi-Hsuan Tsai,
  and Ming-Hsuan Yang.
\newblock Weakly-supervised semantic segmentation via sub-category exploration.
\newblock In {\em {IEEE Conf. Comput. Vis. Pattern Recog.}}, pages 8991--9000,
  2020.

\bibitem{chen2021end}
Jianjun Chen, Shancheng Fang, Hongtao Xie, Zheng-Jun Zha, Yue Hu, and Jianlong
  Tan.
\newblock End-to-end boundary exploration for weakly-supervised semantic
  segmentation.
\newblock In {\em {ACM Int. Conf. Multimedia}}, pages 2381--2390, 2021.

\bibitem{chen2020weakly}
Liyi Chen, Weiwei Wu, Chenchen Fu, Xiao Han, and Yuntao Zhang.
\newblock Weakly supervised semantic segmentation with boundary exploration.
\newblock In {\em {Eur. Conf. Comput. Vis.}}, pages 347--362. Springer, 2020.

\bibitem{chen2014semantic}
Liang-Chieh Chen, George Papandreou, Iasonas Kokkinos, Kevin Murphy, and Alan~L
  Yuille.
\newblock Semantic image segmentation with deep convolutional nets and fully
  connected crfs.
\newblock In {\em {Int. Conf. Learn. Represent.}}, 2015.

\bibitem{chen2017deeplab}
Liang-Chieh Chen, George Papandreou, Iasonas Kokkinos, Kevin Murphy, and Alan~L
  Yuille.
\newblock Deeplab: Semantic image segmentation with deep convolutional nets,
  atrous convolution, and fully connected crfs.
\newblock {\em {IEEE Trans. Pattern Anal. Mach. Intell.}}, 40(4):834--848,
  2017.

\bibitem{choe2020attention}
Junsuk Choe, Seungho Lee, and Hyunjung Shim.
\newblock Attention-based dropout layer for weakly supervised single object
  localization and semantic segmentation.
\newblock {\em {IEEE Trans. Pattern Anal. Mach. Intell.}}, 2020.

\bibitem{dai2015boxsup}
Jifeng Dai, Kaiming He, and Jian Sun.
\newblock Boxsup: Exploiting bounding boxes to supervise convolutional networks
  for semantic segmentation.
\newblock In {\em {Int. Conf. Comput. Vis.}}, pages 1635--1643, 2015.

\bibitem{deng2009large}
Jia Deng, Wei Dong, Richard Socher, Li-Jia Li, Kai Li, and Li Fei-Fei.
\newblock Imagenet: A large-scale hierarchical image database.
\newblock In {\em {IEEE Conf. Comput. Vis. Pattern Recog.}}, pages 248--255,
  2009.

\bibitem{dosovitskiy2020image}
Alexey Dosovitskiy, Lucas Beyer, Alexander Kolesnikov, Dirk Weissenborn,
  Xiaohua Zhai, Thomas Unterthiner, Mostafa Dehghani, Matthias Minderer, Georg
  Heigold, Sylvain Gelly, et~al.
\newblock An image is worth 16x16 words: Transformers for image recognition at
  scale.
\newblock In {\em {Int. Conf. Learn. Represent.}}, 2021.

\bibitem{fan2020learning}
Junsong Fan, Zhaoxiang Zhang, Chunfeng Song, and Tieniu Tan.
\newblock Learning integral objects with intra-class discriminator for
  weakly-supervised semantic segmentation.
\newblock In {\em {IEEE Conf. Comput. Vis. Pattern Recog.}}, pages 4283--4292,
  2020.

\bibitem{fan2020employing}
Junsong Fan, Zhaoxiang Zhang, and Tieniu Tan.
\newblock Employing multi-estimations for weakly-supervised semantic
  segmentation.
\newblock In {\em {Eur. Conf. Comput. Vis.}}, pages 332--348. Springer, 2020.

\bibitem{furlanello2018born}
Tommaso Furlanello, Zachary Lipton, Michael Tschannen, Laurent Itti, and Anima
  Anandkumar.
\newblock Born again neural networks.
\newblock In {\em {Int. Conf. Mach. Learn.}}, pages 1607--1616, 2018.

\bibitem{hariharan2011semantic}
Bharath Hariharan, Pablo Arbel{\'a}ez, Lubomir Bourdev, Subhransu Maji, and
  Jitendra Malik.
\newblock Semantic contours from inverse detectors.
\newblock In {\em {Int. Conf. Comput. Vis.}}, pages 991--998, 2011.

\bibitem{he2016deep}
Kaiming He, Xiangyu Zhang, Shaoqing Ren, and Jian Sun.
\newblock Deep residual learning for image recognition.
\newblock In {\em {IEEE Conf. Comput. Vis. Pattern Recog.}}, pages 770--778,
  2016.

\bibitem{he2019knowledge}
Tong He, Chunhua Shen, Zhi Tian, Dong Gong, Changming Sun, and Youliang Yan.
\newblock Knowledge adaptation for efficient semantic segmentation.
\newblock In {\em {IEEE Conf. Comput. Vis. Pattern Recog.}}, pages 578--587,
  2019.

\bibitem{hinton2015distilling}
Geoffrey Hinton, Oriol Vinyals, and Jeff Dean.
\newblock Distilling the knowledge in a neural network.
\newblock In {\em {Adv. Neural Inform. Process. Syst. Worksh.}}, 2014.

\bibitem{hou2018self}
Qibin Hou, PengTao Jiang, Yunchao Wei, and Ming-Ming Cheng.
\newblock Self-erasing network for integral object attention.
\newblock In {\em Advances in Neural Information Processing Systems},
  volume~31, pages 549--559, 2018.

\bibitem{hou2016mining}
Qibin Hou, Daniela Massiceti, Puneet~Kumar Dokania, Yunchao Wei, Ming-Ming
  Cheng, and Philip~HS Torr.
\newblock Bottom-up top-down cues for weakly-supervised semantic segmentation.
\newblock In {\em Int. Worksh. on Energy Minimization Methods in Comput. Vis.
  Pattern Recog.}, pages 263--277, 2017.

\bibitem{huang2018weakly}
Zilong Huang, Xinggang Wang, Jiasi Wang, Wenyu Liu, and Jingdong Wang.
\newblock Weakly-supervised semantic segmentation network with deep seeded
  region growing.
\newblock In {\em {IEEE Conf. Comput. Vis. Pattern Recog.}}, pages 7014--7023,
  2018.

\bibitem{jiang2021online}
Peng-Tao Jiang, Ling-Hao Han, Qibin Hou, Ming-Ming Cheng, and Yunchao Wei.
\newblock Online attention accumulation for weakly supervised semantic
  segmentation.
\newblock {\em {IEEE Trans. Pattern Anal. Mach. Intell.}}, 2021.

\bibitem{jiang2019integral}
Peng-Tao Jiang, Qibin Hou, Yang Cao, Ming-Ming Cheng, Yunchao Wei, and Hong-Kai
  Xiong.
\newblock Integral object mining via online attention accumulation.
\newblock In {\em {Int. Conf. Comput. Vis.}}, pages 2070--2079, 2019.

\bibitem{kim2021discriminative}
Beomyoung Kim, Sangeun Han, and Junmo Kim.
\newblock Discriminative region suppression for weakly-supervised semantic
  segmentation.
\newblock In {\em {AAAI Conf. Artif. Intell.}}, pages 1754--1761, 2021.

\bibitem{kolesnikov2016seed}
Alexander Kolesnikov and Christoph~H Lampert.
\newblock Seed, expand and constrain: Three principles for weakly-supervised
  image segmentation.
\newblock In {\em {Eur. Conf. Comput. Vis.}}, pages 695--711, 2016.

\bibitem{kweon2021unlocking}
Hyeokjun Kweon, Sung-Hoon Yoon, Hyeonseong Kim, Daehee Park, and Kuk-Jin Yoon.
\newblock Unlocking the potential of ordinary classifier: Class-specific
  adversarial erasing framework for weakly supervised semantic segmentation.
\newblock In {\em {Int. Conf. Comput. Vis.}}, pages 6994--7003, 2021.

\bibitem{lee2019ficklenet}
Jungbeom Lee, Eunji Kim, Sungmin Lee, Jangho Lee, and Sungroh Yoon.
\newblock Ficklenet: Weakly and semi-supervised semantic image segmentation
  using stochastic inference.
\newblock In {\em {IEEE Conf. Comput. Vis. Pattern Recog.}}, pages 5267--5276,
  2019.

\bibitem{lee2021anti}
Jungbeom Lee, Eunji Kim, and Sungroh Yoon.
\newblock Anti-adversarially manipulated attributions for weakly and
  semi-supervised semantic segmentation.
\newblock In {\em {IEEE Conf. Comput. Vis. Pattern Recog.}}, pages 4071--4080,
  2021.

\bibitem{lee2021railroad}
Seungho Lee, Minhyun Lee, Jongwuk Lee, and Hyunjung Shim.
\newblock Railroad is not a train: Saliency as pseudo-pixel supervision for
  weakly supervised semantic segmentation.
\newblock In {\em {IEEE Conf. Comput. Vis. Pattern Recog.}}, pages 5495--5505,
  2021.

\bibitem{li2018tell}
Kunpeng Li, Ziyan Wu, Kuan-Chuan Peng, Jan Ernst, and Yun Fu.
\newblock Tell me where to look: Guided attention inference network.
\newblock In {\em {IEEE Conf. Comput. Vis. Pattern Recog.}}, pages 9215--9223,
  2018.

\bibitem{li2021group}
Xueyi Li, Tianfei Zhou, Jianwu Li, Yi Zhou, and Zhaoxiang Zhang.
\newblock Group-wise semantic mining for weakly supervised semantic
  segmentation.
\newblock In {\em {AAAI Conf. Artif. Intell.}}, pages 1984--1992, 2021.

\bibitem{li2021pseudo}
Yi Li, Zhanghui Kuang, Liyang Liu, Yimin Chen, and Wayne Zhang.
\newblock Pseudo-mask matters in weakly-supervised semantic segmentation.
\newblock In {\em {Int. Conf. Comput. Vis.}}, pages 6964--6973, 2021.

\bibitem{lin2016scribblesup}
Di Lin, Jifeng Dai, Jiaya Jia, Kaiming He, and Jian Sun.
\newblock Scribblesup: Scribble-supervised convolutional networks for semantic
  segmentation.
\newblock In {\em {IEEE Conf. Comput. Vis. Pattern Recog.}}, pages 3159--3167,
  2016.

\bibitem{lin2016refinenet}
Guosheng Lin, Anton Milan, Chunhua Shen, and Ian Reid.
\newblock Refinenet: Multi-path refinement networks for high-resolution
  semantic segmentation.
\newblock In {\em {IEEE Conf. Comput. Vis. Pattern Recog.}}, pages 1925--1934,
  2017.

\bibitem{liu2019simple}
Jiang-Jiang Liu, Qibin Hou, Ming-Ming Cheng, Jiashi Feng, and Jianmin Jiang.
\newblock A simple pooling-based design for real-time salient object detection.
\newblock In {\em {IEEE Conf. Comput. Vis. Pattern Recog.}}, pages 3917--3926,
  2019.

\bibitem{liu2019structured}
Yifan Liu, Ke Chen, Chris Liu, Zengchang Qin, Zhenbo Luo, and Jingdong Wang.
\newblock Structured knowledge distillation for semantic segmentation.
\newblock In {\em {IEEE Conf. Comput. Vis. Pattern Recog.}}, pages 2604--2613,
  2019.

\bibitem{liu2020leveraging}
Yun Liu, Yu-Huan Wu, Pei-Song Wen, Yu-Jun Shi, Yu Qiu, and Ming-Ming Cheng.
\newblock Leveraging instance-, image-and dataset-level information for weakly
  supervised instance segmentation.
\newblock {\em {IEEE Trans. Pattern Anal. Mach. Intell.}}, 2020.

\bibitem{long2015fully}
Jonathan Long, Evan Shelhamer, and Trevor Darrell.
\newblock Fully convolutional networks for semantic segmentation.
\newblock In {\em {IEEE Conf. Comput. Vis. Pattern Recog.}}, pages 3431--3440,
  2015.

\bibitem{papandreou2015weakly}
George Papandreou, Liang-Chieh Chen, Kevin~P Murphy, and Alan~L Yuille.
\newblock Weakly-and semi-supervised learning of a deep convolutional network
  for semantic image segmentation.
\newblock In {\em {Int. Conf. Comput. Vis.}}, pages 1742--1750, 2015.

\bibitem{pathak2014fully}
Deepak Pathak, Evan Shelhamer, Jonathan Long, and Trevor Darrell.
\newblock Fully convolutional multi-class multiple instance learning.
\newblock In {\em {Int. Conf. Learn. Represent.}}, 2015.

\bibitem{pinheiro2015image}
Pedro~O Pinheiro and Ronan Collobert.
\newblock From image-level to pixel-level labeling with convolutional networks.
\newblock In {\em {IEEE Conf. Comput. Vis. Pattern Recog.}}, pages 1713--1721,
  2015.

\bibitem{shimoda2019self}
Wataru Shimoda and Keiji Yanai.
\newblock Self-supervised difference detection for weakly-supervised semantic
  segmentation.
\newblock In {\em {Int. Conf. Comput. Vis.}}, pages 5208--5217, 2019.

\bibitem{simonyan2014very}
Karen Simonyan and Andrew Zisserman.
\newblock Very deep convolutional networks for large-scale image recognition.
\newblock In {\em {Int. Conf. Learn. Represent.}}, 2015.

\bibitem{su2021context}
Yukun Su, Ruizhou Sun, Guosheng Lin, and Qingyao Wu.
\newblock Context decoupling augmentation for weakly supervised semantic
  segmentation.
\newblock In {\em {Int. Conf. Comput. Vis.}}, 2021.

\bibitem{sun2020mining}
Guolei Sun, Wenguan Wang, Jifeng Dai, and Luc Van~Gool.
\newblock Mining cross-image semantics for weakly supervised semantic
  segmentation.
\newblock In {\em {Eur. Conf. Comput. Vis.}}, pages 347--365, 2020.

\bibitem{sun2021ecs}
Kunyang Sun, Haoqing Shi, Zhengming Zhang, and Yongming Huang.
\newblock Ecs-net: Improving weakly supervised semantic segmentation by using
  connections between class activation maps.
\newblock In {\em {Int. Conf. Comput. Vis.}}, pages 7283--7292, 2021.

\bibitem{vernaza2017learning}
Paul Vernaza and Manmohan Chandraker.
\newblock Learning random-walk label propagation for weakly-supervised semantic
  segmentation.
\newblock In {\em {IEEE Conf. Comput. Vis. Pattern Recog.}}, pages 7158--7166,
  2017.

\bibitem{wang2018weakly}
Xiang Wang, Shaodi You, Xi Li, and Huimin Ma.
\newblock Weakly-supervised semantic segmentation by iteratively mining common
  object features.
\newblock In {\em {IEEE Conf. Comput. Vis. Pattern Recog.}}, pages 1354--1362,
  2018.

\bibitem{wang2020self}
Yude Wang, Jie Zhang, Meina Kan, Shiguang Shan, and Xilin Chen.
\newblock Self-supervised equivariant attention mechanism for weakly supervised
  semantic segmentation.
\newblock In {\em {IEEE Conf. Comput. Vis. Pattern Recog.}}, pages
  12275--12284, 2020.

\bibitem{wei2017object}
Yunchao Wei, Jiashi Feng, Xiaodan Liang, Ming-Ming Cheng, Yao Zhao, and
  Shuicheng Yan.
\newblock Object region mining with adversarial erasing: A simple
  classification to semantic segmentation approach.
\newblock In {\em {IEEE Conf. Comput. Vis. Pattern Recog.}}, pages 1568--1576,
  2017.

\bibitem{wei2018revisiting}
Yunchao Wei, Huaxin Xiao, Honghui Shi, Zequn Jie, Jiashi Feng, and Thomas~S
  Huang.
\newblock Revisiting dilated convolution: A simple approach for weakly-and
  semi-supervised semantic segmentation.
\newblock In {\em {IEEE Conf. Comput. Vis. Pattern Recog.}}, pages 7268--7277,
  2018.

\bibitem{wu2021embedded}
Tong Wu, Junshi Huang, Guangyu Gao, Xiaoming Wei, Xiaolin Wei, Xuan Luo, and
  Chi~Harold Liu.
\newblock Embedded discriminative attention mechanism for weakly supervised
  semantic segmentation.
\newblock In {\em {IEEE Conf. Comput. Vis. Pattern Recog.}}, pages
  16765--16774, 2021.

\bibitem{wu2019wider}
Zifeng Wu, Chunhua Shen, and Anton Van Den~Hengel.
\newblock Wider or deeper: Revisiting the resnet model for visual recognition.
\newblock {\em {Pattern Recogn.}}, 90:119--133, 2019.

\bibitem{xu2021leveraging}
Lian Xu, Wanli Ouyang, Mohammed Bennamoun, Farid Boussaid, Ferdous Sohel, and
  Dan Xu.
\newblock Leveraging auxiliary tasks with affinity learning for weakly
  supervised semantic segmentation.
\newblock In {\em {Int. Conf. Comput. Vis.}}, pages 6984--6993, 2021.

\bibitem{yao2020saliency}
Qi Yao and Xiaojin Gong.
\newblock Saliency guided self-attention network for weakly and semi-supervised
  semantic segmentation.
\newblock {\em IEEE Access}, 8:14413--14423, 2020.

\bibitem{yao2021non}
Yazhou Yao, Tao Chen, Guo-Sen Xie, Chuanyi Zhang, Fumin Shen, Qi Wu, Zhenmin
  Tang, and Jian Zhang.
\newblock Non-salient region object mining for weakly supervised semantic
  segmentation.
\newblock In {\em {IEEE Conf. Comput. Vis. Pattern Recog.}}, pages 2623--2632,
  2021.

\bibitem{yuan2021tokens}
Li Yuan, Yunpeng Chen, Tao Wang, Weihao Yu, Yujun Shi, Zi-Hang Jiang,
  Francis~EH Tay, Jiashi Feng, and Shuicheng Yan.
\newblock Tokens-to-token vit: Training vision transformers from scratch on
  imagenet.
\newblock In {\em {Int. Conf. Comput. Vis.}}, pages 558--567, 2021.

\bibitem{yuan2020revisiting}
Li Yuan, Francis~EH Tay, Guilin Li, Tao Wang, and Jiashi Feng.
\newblock Revisiting knowledge distillation via label smoothing regularization.
\newblock In {\em {IEEE Conf. Comput. Vis. Pattern Recog.}}, pages 3903--3911,
  2020.

\bibitem{zhang2020reliability}
Bingfeng Zhang, Jimin Xiao, Yunchao Wei, Mingjie Sun, and Kaizhu Huang.
\newblock Reliability does matter: An end-to-end weakly supervised semantic
  segmentation approach.
\newblock In {\em {AAAI Conf. Artif. Intell.}}, pages 12765--12772, 2020.

\bibitem{zhang2018adversarial}
Xiaolin Zhang, Yunchao Wei, Jiashi Feng, Yi Yang, and Thomas~S Huang.
\newblock Adversarial complementary learning for weakly supervised object
  localization.
\newblock In {\em {IEEE Conf. Comput. Vis. Pattern Recog.}}, pages 1325--1334,
  2018.

\bibitem{zhao2016pyramid}
Hengshuang Zhao, Jianping Shi, Xiaojuan Qi, Xiaogang Wang, and Jiaya Jia.
\newblock Pyramid scene parsing network.
\newblock In {\em {IEEE Conf. Comput. Vis. Pattern Recog.}}, pages 2881--2890,
  2017.

\bibitem{zhou2016learning}
Bolei Zhou, Aditya Khosla, Agata Lapedriza, Aude Oliva, and Antonio Torralba.
\newblock Learning deep features for discriminative localization.
\newblock In {\em {IEEE Conf. Comput. Vis. Pattern Recog.}}, pages 2921--2929,
  2016.

\bibitem{zhou2020multi}
Kuangqi Zhou, Qibin Hou, Zun Li, and Jiashi Feng.
\newblock Multi-miner: Object-adaptive region mining for weakly-supervised
  semantic segmentation.
\newblock {\em arXiv preprint arXiv:2006.07834}, 2020.

\end{thebibliography}
}

\end{document}